\documentclass[conference]{IEEEtran}
\IEEEoverridecommandlockouts
\usepackage{cite}
\usepackage{amsmath,amssymb,amsfonts}
\usepackage{algorithmic}
\usepackage{graphicx}
\usepackage{textcomp}
\usepackage{xcolor}
\usepackage{comment}

\renewcommand\IEEEkeywordsname{Keywords}

\usepackage{fancyhdr}

\usepackage{caption}
\usepackage{siunitx}
\usepackage{booktabs}
\usepackage{tabularx}
\newcolumntype{Y}{>{\centering\arraybackslash}X}

\def\BibTeX{{\rm B\kern-.05em{\sc i\kern-.025em b}\kern-.08em
    T\kern-.1667em\lower.7ex\hbox{E}\kern-.125emX}}
\begin{document}

\title{Named Entity Recognition for Monitoring Plant Health Threats in Tweets: a ChouBERT Approach\\

\thanks{Shufan Jiang is also affiliated to Institut Supérieur d'Electronique de Paris, France}
}

\makeatletter 
\newcommand{\linebreakand}{%
  \end{@IEEEauthorhalign}
  \hfill\mbox{}\par
  \mbox{}\hfill\begin{@IEEEauthorhalign}
}
\makeatother 

\author{
\IEEEauthorblockN{Shufan Jiang\thanks{\IEEEauthorrefmark{1}Corresponding author}\IEEEauthorrefmark{1}}
\IEEEauthorblockA{\textit{Centre de Recherche en STIC} \\
\textit{University of Reims Champagne-Ardenne}\\
Reims, France \\
jiang.chou.fan@gmail.com}

\and
\IEEEauthorblockN{Rafael Angarita}
\IEEEauthorblockA{
\textit{Institut Supérieur d'Electronique de Paris}\\
Issy-les-Moulineaux, France \\
rangarit@parisnanterre.fr}

\and
\IEEEauthorblockN{St\'ephane Cormier}
\IEEEauthorblockA{\textit{Centre de Recherche en STIC} \\
\textit{University of Reims Champagne-Ardenne}\\
Reims, France \\
stephane.cormier@univ-reims.fr}

\linebreakand 
\IEEEauthorblockN{Francis Rousseaux}
\IEEEauthorblockA{\textit{Centre de Recherche en STIC} \\
\textit{University of Reims Champagne-Ardenne}\\
Reims, France \\
francis.rousseaux@univ-reims.fr}

}

\maketitle

\thispagestyle{fancy}
\fancyhead{} 
\fancyfoot{}
\lhead{IEEE 6th International Conference on Universal Village · UV2022 · Session TS5-B-9} 
\lfoot{\fontsize{8}{10} \selectfont 978-1-6654-7477-1/22/\$31.00 \copyright2022 IEEE} 
\rhead{}
\cfoot{} 
\rfoot{}

\begin{abstract}
An important application scenario of precision agriculture is detecting and measuring crop health threats using sensors and data analysis techniques. However, the textual data are still under-explored among the existing solutions due to the lack of labelled data and fine-grained semantic resources. Recent research suggests that the increasing connectivity of farmers and the emergence of online farming communities make social media like Twitter a participatory platform for detecting unfamiliar plant health events if we can extract essential information from unstructured textual data. ChouBERT is a French pre-trained language model that can identify Tweets concerning observations of plant health issues with generalizability on unseen natural hazards. This paper tackles the lack of labelled data by further studying ChouBERT's know-how on token-level annotation tasks over small labeled sets.
\end{abstract}

\hfill  


In the agricultural domain, textual data like periodic plant health bulletins, farmers' notes and posts on the Internet contains rich information about the observation of crops and farmers' experiences dedicated to specific environmental and historical conditions.
Followed by our work ChouBERT~\cite{jiang2022choubert}, we aim to continue the development of an epidemiological surveillance system based on tweets. Unlike the existing Twitter-based surveillance systems that focus on specific issues~\cite{welvaert_limits_2017}, ChouBERT's text classifier processes any tweets concerning plant health. Now that we can find the plant health observation information, our next step will be automatically identifying the natural hazards and impacted crops in such text. In information extraction (IE), this work can be done by two tasks: Named Entity Recognition (NER) and  Named Entity Linking (NEL). Named entities are phrases that contain the names of persons, organizations, locations, numerical expressions, in our case, disease, pest or crop. The entities and the relations among them are the essential elements of formal knowledge graph construction. We can use the knowledge graph to index and integrate information from heterogeneous documents~\cite{Jiang2020}. Given a text, we split it into a sequence of tokens $S = ( w_1, w_2, ..., w_n )$, the goal of NER is then to identify whether a subsequence $S' = (w_k, ..., w_l ) ,(1 \leq k \leq l \leq n)$ is an entity.  Then the goal of NEL is to assign the identified phrase to a unique concept in existing knowledge graphs. 
A recent survey about limitations of IE~\cite{limitation_ie} summarises the challenges to text-based event extraction from tweets as (a) the ambiguity of representation, (b) noisy data and (c) lack of training data. The text classification results of ChouBERT~\cite{jiang2022choubert} prove its capacity to represent plant health-related information and filter out noises among the tweets. This paper addresses the lack of training data by examining if ChouBERT-based NER improves the detection of named entities of natural hazards in tweets with small sizes of labelled data.

\section{Related Work}

\subsection{NER for Epidemic Monitoring}
The authors of~\cite{limitation_ie} categorize NER technologies into rule-based, machine-learning and hybrid approaches. In this sense, we review the available resources to analyze the feasibility of these approaches towards monitoring plant health threats on Twitter.  

Rule-based approaches depend on grammar rules and dictionaries to describe human languages for computers. 
To achieve our domain-specific classification tasks, we generalize grammar rules and dictionaries to machine-comprehensive knowledge about plant health, like a list of known diseases of an apple variety or a list of environmental conditions that favour the germination of fungal spores. An example of domain non-specific entity extraction with rule-based systems is PADI-web~\cite{padi-web2020}, a French epidemiological animal health surveillance platform. PADI-web uses rules and gazetteers to extract locations and dates. Considering domain-specific knowledge graphs about plant health, FrenchCropUsage (FCU)~\cite{FCU_2021} is a French thesaurus of crops organized by usage; the project VESPA~\cite{turenne_open} produce a list of disease names and a list of pest names to index French Plant Health Bulletins using hand-crafted rules. These existing dictionaries in the French plant health domain enable information retrieval by the occurrences of terms but cannot support the disambiguation in NER nor give insights into hybrid approaches.

Machine learning approaches model a language as a probability distribution over sequences of words~\cite{jurafsky2018speech}. Then the machine can learn to classify the data points based on their similarity if the modelling extracts representative features in the text. Thus we can divide machine learning-based NER approaches into two parts: (1) converting text into vectors (aka feature engineering) and (2) applying classification algorithms to the vectors. A popular supervised classification algorithm is Conditional Random Fields (CRF)~\cite{CRF}, such as~\cite{yu_adversarial_2020, CRF_NER}. However, supervised approaches demand a large quantity of labelled data~\cite{limitation_ie}. Semi-supervised learning can improve the classifier's performance by benefiting from unlabelled data. The authors of~\cite{yu_adversarial_2020} propose combining two semi-supervised approaches for identifying medical concepts and annotation inconsistency: using pre-trained encoders BERT or clinical-domain BioWordVec for vectorizing the text and applying adversarial learning above CRF classifiers. BioWordVec gives static embedding to each token, and BERT gives contextualized embedding to each token. The results in ~\cite{yu_adversarial_2020} show that the differences between BERT-based and BioWordVec-based  are always much more significant than the difference between these models and their GAN versions -- with GAN or not, BERT-based outperforms BioWordVec-based, which emphasises the significance of feature engineering. Thus, we decide first to investigate if such representation improves the detection of named entities of natural hazards in tweets with small sizes of labelled data. We will give detailed introduction to pre-trained language models in the next subsection.

\subsection{Pre-trained Language Models}
Pre-trained language models (PLM) are deep neural networks with pre-trained weights that vectorize word sequences.
Such vectorial representations yield cutting-edge outcomes on NLP tasks like text classification, text clustering, question-answering, and information extraction.
PLMs suggest an objective engineering paradigm for NLP: pre-training language models to extract contextualized characteristics from text, followed by fine-tuning with task-specific objective functions~\cite{liu_pre-train_2021}.
Bidirectional encoder representations based on transformers (BERT)~\cite{devlin_bert_2019} is a PLM developed by Google that has significantly advanced this area since it was first announced in 2018.
BERT is pre-trained in two stages: first, a self-supervised task in which the Masked Language Model (MLM) has to discover words that have been masked in a text; and second, a supervised task in which the model has to determine whether a sentence B is the continuation of a sentence A (Next-Sentence Prediction, NSP). 
The pre-training produces at the end 12 stacked encoders, which take a sequence of tokens as input and calculate a fix-length vector for each token. 
These vectors' dimensions correspond to how much consideration a token should give to the other tokens.
The CamemBERT model~\cite{martin_camembert_2020}, one of the French variants of BERT, is based on the same architecture as BERT and was trained using MLM exclusively on a French corpus.
ChouBERT~\cite{jiang2022choubert} further pre-trains a CamemBERT-base checkpoint using MLM over French Plant Health Bulletins and Tweets mentioning natural hazards in French. The pre-training of ChouBERT improves performance in identifying plant health issues on Twitter with relatively small labelled data.

\section{Methodology}

\subsection{Dataset for NER}
   \subsubsection{Annotation}
     We use disease names and pest names produced by~\cite{turenne_open} to collect tweets using Twitter's full-archive search API. For each hazard, we sample up to 5 tweets. To include as many different hazards in the labelled set as possible, we do not filter these tweets with any observation classifiers from the previous session. We manually annotate 1028 tweets with INCEpTION~\cite{INCEpTION2018} and export the labelled data with IOB2 tagging, where B- prefixed tag denotes the first term of every named entity, I- prefixed tag indicates any non-initial term and the O tag means that the token is outside any target entities. As we aim to identify diseases and pests, there are five different tags for each token: B-maladie (the beginning of any disease), I-maladie, B-ravageur (the beginning of any pest), I-ravageur and O. 
     As the CamemBERT tokeniser could break a word into several tokens (wordpieces), we tokenise all the labelled data and label the wordpieces of a word with its IOB2 label. So, for example, the disease entity ``phoma du colza'' with its original IOB2 tagging [B-Maladie, I-Maladie, I-Maladie], is tokenised intro ``\_pho ma \_du \_colza'', then the labels for theses wordpieces are [B-Maladie, B-Maladie, I-Maladie, I-Maladie]. All our evaluation metrics are then calculated based on the predictions for each wordpiece.
 
    \subsubsection{Train-validation-test Split}
    As ChouBERT~\cite{jiang2022choubert} claims its advantage in classifying unseen natural hazards and evaluates such capacity with the polysemous term ``taupin'', we wonder if ChouBERT representation helps to detect unseen and ambiguous hazard names. Thus, we select a list of hazard names in Table~\ref{tab:validationlist} and use all the tweets containing such terms to make a test set of 207 tweets. Then we sampled 640 tweets to make 5-fold train-validation sets for cross-validation. At last, we append the rest 181 tweets to each of the five validation sets. So in the validation set, there are seen and unseen hazards, while in the test set, there are only unseen hazards. 

    \begin{table}[hbt!]
        \begin{center}           
        \caption{List of Hazards in the Test Set and Their Meaning(s)}
        \label{tab:validationlist}
        \tabcolsep = 1\tabcolsep
            \begin{tabular}{|c|>{\centering\arraybackslash}m{.5\linewidth}|}
            \hline
                \textbf{Hazard} & \textbf{Meaning(s)} \\ \hline
                oïdium or oidium & Powdery mildew, a fungal disease \\ \hline
                teigne & 1. an insect pest, e.g. teigne du poireaux 2.  Dermatophytosis, a fungal infection of human skin 3. the title of the monarch of the pre-colonial Kingdom of Baol  \\ \hline
                rouille & 1. rust (fungus), plant diseases 2. rust ( iron oxide ) 3. a sauce in French cuisine \\ \hline
                mosaïque or mosaique & 1. Mosaic, decorative art style 2. Mosaic virus \\ \hline
                pourriture & 1. rottenness, fungal diseases 2. political corruption \\ \hline
                taupe & 1. Mole, a small fossorial mammal 2. a family name \\ \hline
                taupin & 1. wireworm, an insect pest 2. a family name 3. an undergraduate student from a French scientific preparatory class \\ \hline
                mouche & 1. flies, insect pests 2. Bateaux Mouches, excursion boats along the river Seine \\ \hline
                tipule & tipula, a very large insect genus in the fly family Tipulidae \\ \hline
                cousin & 1. homonym to ``tipule'' 2. cousin, a type of familial relationship  \\ \hline
            \end{tabular}
        \end{center}
    \end{table}

\subsection{Experimental Setup}
We use the implementation for token classification \textit{CamemebertForTokenClassification} in the transformers package~\cite{wolf-etal-2020-transformers}. 
It loads a pre-trained CamemeBERT model and adds a linear layer on top of the token representation output. In this study, we load the out-of-box CamemBERT-base model, the ChouBERT model pre-trained for 16 epochs (denoted as ChouBERT-16), and the ChouBERT model pre-trained for 32 epochs (denoted as ChouBERT-32). Then we train the linear layer to predict the probability of a token's representation matching one of the five labels. As~\cite{croce_gan-bert_2020} claims:
\begin{quote}
   ``… the quality of BERT fine-tuned over less than 200 annotated instances shows significant drops, especially in classification tasks involving many categories.''
\end{quote}
We aim to see if the pre-training of ChouBERT improves the NER performance when there are less than 200 annotated instances. Thus, for each group of our train-validation dataset, we sampled 16, 32, 64, 128, 256, and 512 instances from the train set to train the NER classifier and note the precision, recall and F1-score over the validation set and the test set. We set the maximum sequence length of the model to 128. To fix the total training steps, we set the batch size to (size of the data for training / 16). Thus the number of training steps in each epoch is fixed at 16. We run all our experiments with a fixed learning rate of $5e-5$ for 20 epochs.

\subsection{Results and Evaluation}
\begin{figure}[hbt!]
\begin{center}
\includegraphics[width=\linewidth]{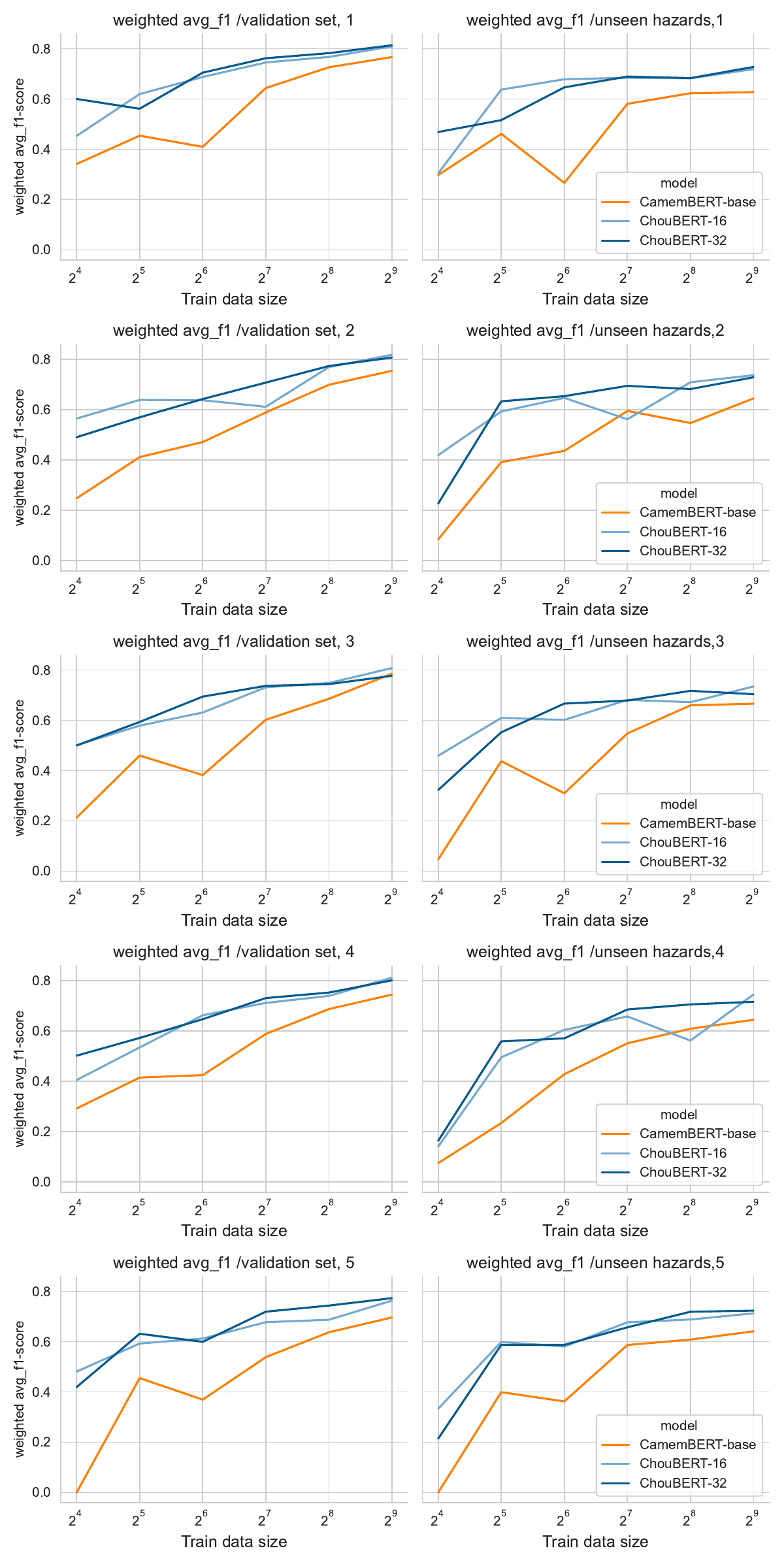}
\end{center}
\caption{ Weighted Average F1 of NER on Validation Set and Test Set}
\label{fig:weight_avg_f1_NER}
\end{figure}
\begin{figure}[hbt!]
\begin{center}
\includegraphics[width=\linewidth]{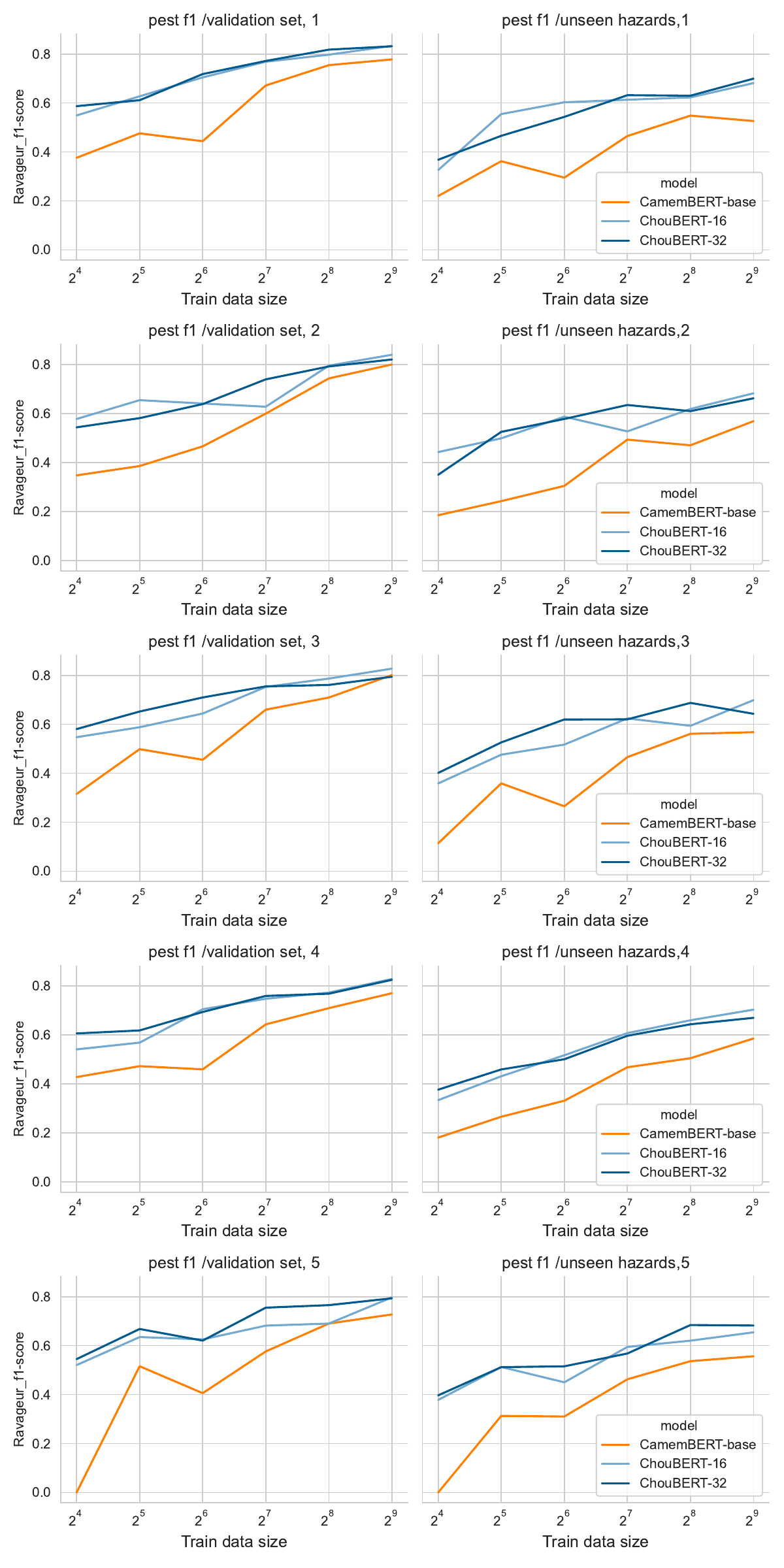}%
\end{center}
\caption{ F1 of Pest NER on Validation Set and Test Set}
\label{fig:pest_f1_NER}
\end{figure}
\begin{figure}[hbt!]
\begin{center}
\includegraphics[width=\linewidth]{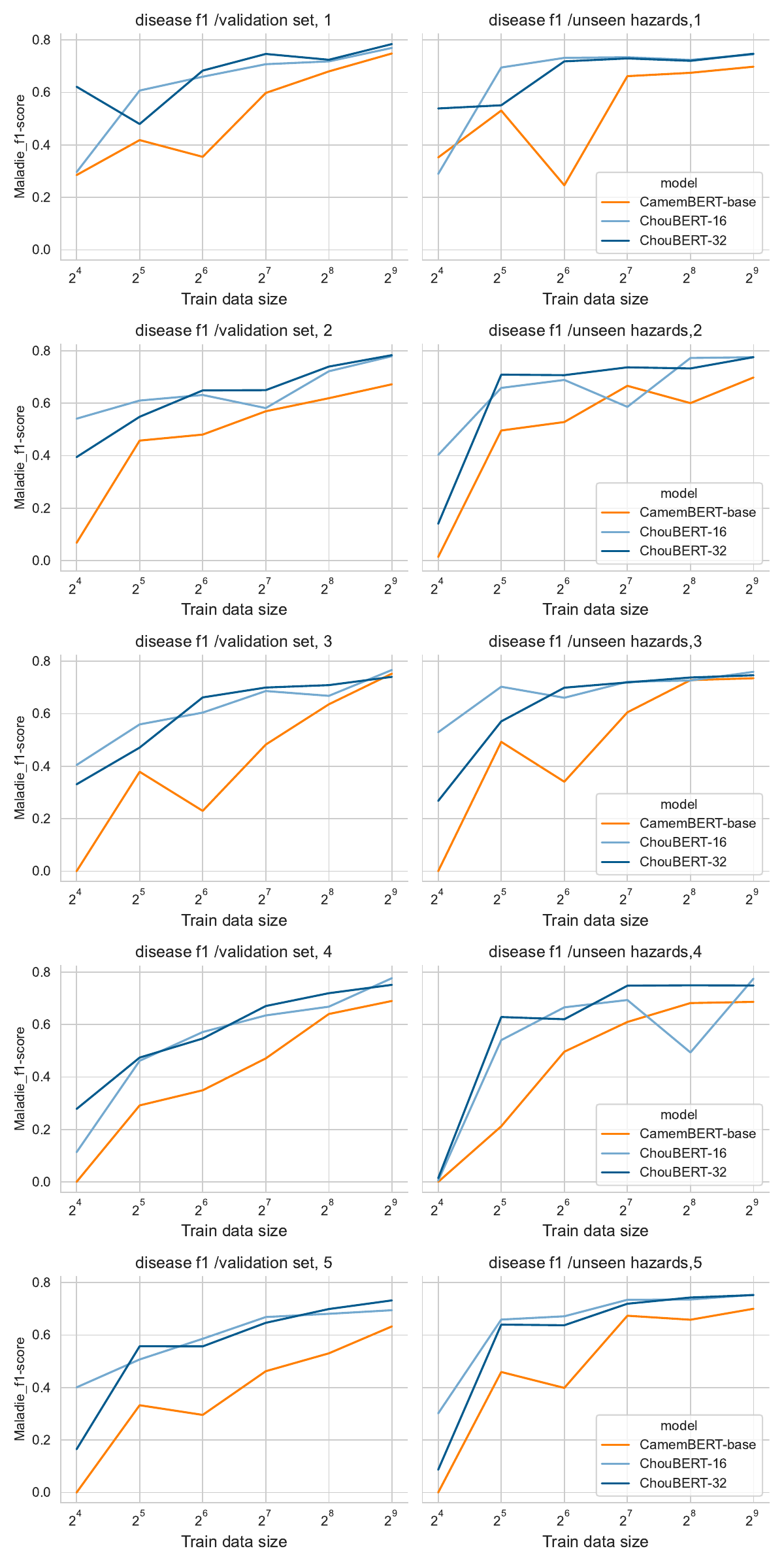}%
\end{center}
\caption{ F1 of Disease NER on Validation Set and Test Set}
\label{fig:disease_f1_NER}
\end{figure}
We illustrate the evolution of the average weighted F1 score of the NER classification with different training data sizes in Fig.~\ref{fig:weight_avg_f1_NER}, the F1 scores of Pest class in Fig.~\ref{fig:pest_f1_NER} and the F1 scores of Disease class in Fig.~\ref{fig:disease_f1_NER}. We note the best F1 scores of 15 training epochs for each combination of unlabelled size, train size, and PLM. Remarkably, both ChouBERT-16 and ChouBERT-32 outperform CamemBERT-base initially, where we train the classifiers with only 16 labelled tweets. With less than 256 labelled tweets, there is always a significant distance between the CamemBERT-base classifier and the ChouBERT classifiers. Such distance tends to reduce when there is more data, but in most experiments, ChouBERT classifiers are better on the validation and unseen hazards sets. These results being coherent to those in the~\cite{jiang2022choubert}, proves that the pre-training of ChouBERT improves different downstream NLP tasks in the plant health domain.

\subsection{Threats to Validity}
As huggingface's implementation~\cite{wolf-etal-2020-transformers} of NER applies an independent linear classifier on the top of each token representation, the training does not explicitly ensure the coherence between neighbour labels. Thus, it is rare but possible that a token is classified as a non-beginning entity token like I-maladie without the token before it being classified as B-maladie or I-maladie. We evaluate the classifier's performance based on the prediction of each token. Nevertheless, we only combine coherent sequences when we yield the final entity annotations from the classified wordpieces.  

\section{Conclusion}
In this study, we validate ChouBERT's capacity to detect plant health-related entities over small quantities of labelled data and its generalizability to unseen and ambiguous natural hazards. Our work opens up the fast integration of heterogeneous textual documents in the French plant health context with ChouBERT. Future directions include developing other IE tasks with ChouBERT, like entity-linking and relation extraction; annotating more fine-grained information like the symptoms and the developing stages of hazards on crops; optimizing the model with knowledge distillation; and investigating hybrid approaches with newly developed knowledge graphs.

\bibliographystyle{IEEEtran}
\bibliography{IEEEabrv, IEEEexample}

\end{document}